\documentclass[10pt,twoside]{article}

\usepackage{times}
\usepackage{amsmath}

\usepackage[latin1]{inputenc}
\usepackage[french]{babel}
\usepackage[dvips]{epsfig}
\usepackage{fancyhdr}

\makeatother
\renewenvironment{abstract}{%
      \list{}{\leftmargin=1cm
      \labelwidth=1cm
      \listparindent=1cm
      \itemindent\listparindent
      \rightmargin\leftmargin}\item[\hskip\labelsep
                                    \bfseries Résumé.]}
    {\endlist}
\makeatletter

\newcommand{\argmax}{\operatornamewithlimits{argmax}}

\begin{document}
\thispagestyle{empty}

\fancyhead{}
\fancyfoot{}
\fancyhead[LE]{
%
%
Fusion de classifieurs pour la classification d'images sonar
%
}
\fancyhead[RO]{
%
Arnaud MARTIN
%
}
\fancyfoot[LE,RO]{RNTI - 1}

\begin{center}
{\Large\bf
%
Fusion de classifieurs pour la classification d'images sonar
%
}
~\\~\\
%
Arnaud MARTIN\\
%
~\\
%
%
ENSIETA / E$^3$I$^2$, EA3876\\
  2, rue François Verny, 29806 Brest cedex 9\\
	Arnaud.Martin@ensieta.fr\\
	http://www.ensieta.fr/e3i2\\
\end{center}
%
%
\begin{abstract}
Nous présentons dans ce papier des approches de fusion d'informations haut niveau applicables pour des données numériques ou des données symboliques. Nous étudions l'intérêt des telles approches particulièrement pour la fusion de classifieurs. Une étude comparative est présentée dans le cadre de la caractérisation des fonds marins à partir d'images sonar. Reconnaître le type de sédiments sur des images sonar est un problème difficile en soi en partie à cause de la complexité des données. Nous comparons les approches de fusion d'informations haut niveau et montrons le gain obtenu.
\end{abstract}
%

%
\section{Introduction}
La fusion d'informations est apparue afin de gérer des quantités très importantes de données multisources dans le domaine militaire. Depuis quelques années des méthodes de fusion ont été adaptées et développées pour des applications en traitement du signal. Plusieurs sens sont donnés à la fusion d'informations, nous reprenons ici la définition proposée par (Bloch 2003)~: La fusion d'informations consiste à combiner des informations issues de plusieurs sources afin d'aider à la prise de décision.

Nous ne cherchons pas ici à réduire les redondances contenues dans les informations issues de plusieurs sources, mais au contraire à en tenir compte afin d'améliorer la prise de décision. De même nous cherchons à modéliser au mieux les différentes imperfections des données (imprécisions, incertitudes, conflit, ambiguïté, incomplétude, fiabilité des sources, ...) non pas pour les supprimer, mais encore pour l'aide à la décision. 

Différents niveaux de fusion ont été proposé dans la littérature. Ce qui est communément retenu, est une division en trois niveaux (Dasarathy 1997), celui des données (ou bas niveau), celui des caractéristiques ({\it i.e.} des paramètres extraits) (ou fusion de niveau intermédiaire) et celui des décisions (ou fusion de haut niveau).

Le choix du niveau de fusion doit se faire en fonction des données disponibles et de l'architecture de la fusion retenue (centralisée, distribuée, ...) qui sont liées à l'application recherchée. Ainsi, nous pouvons chercher à fusionner des informations issues de différents capteurs tels que des radars de fréquences différentes afin d'estimer au mieux la réflexion d'une cible. Dans ce cas une approche de fusion bas niveau sera préférable.

Dans ce papier, nous considérons une application dans le cadre de la classification. Plusieurs classifieurs peuvent fournir une information sur la classe de l'objet observé. Ainsi, nous retenons des approches de fusion haut niveau pour résoudre un tel problème. Les données exprimant une décision peuvent être de type numérique (tel que les sorties des classifieurs) ou symbolique (tel que les classes décidées par les classifieurs exprimées sous forme de symboles). Nous présentons ici une étude comparative des méthodes de fusion haut niveau dans le cadre de la classification d'images sonar. 

Les images sonar sont caractérisées par un grand nombre d'imperfections telles que l'incertitude du milieu, les imprécisions des mesures et de reconstruction de l'image. C'est pourquoi, nous cherchons ici à extraire des paramètres de textures sur ces images, en considérant que la physique du problème a été au mieux prise en compte lors de l'étape de reconstruction. Nous retenons quatre jeux de paramètres extraits selon quatre méthodes différentes de traitement d'images. Les images sonar sont ensuite classées par quatre perceptrons multicouche, chacun considérant un des quatre jeux de paramètres.

Nous présentons tout d'abord trois grandes classes de méthodes de fusion haut niveau, les approches par vote, celles issues de la théorie des possibilités et celles issues de la théorie des croyances. Nous exposons ensuite la complexité des images sonar pour leur classification automatique, ainsi que les quatre méthodes retenues pour l'extraction de paramètres de texture. Enfin nous présentons une évaluation comparative des méthodes de fusion d'informations haut niveau selon la configuration retenue pour la classification d'images sonar.

\section{Méthodes de fusion d'informations}
Nous présentons ici trois cadres théoriques de fusion d'informations haut niveau, le principe du vote, la théorie des possibilités et la théorie des croyances. Nous considérons le problème de la fusion de $m$ sources $S_j$ afin de déterminer une des $n$ classes $C_i$ possibles.

\subsection{Principe du vote}
Le principe du vote est la méthode de fusion d'informations la plus simple à mettre en {\oe}uvre. Plus qu'une approche de fusion, le principe du vote est une méthode de combinaison particulièrement adaptée aux décisions de type symbolique. Notons $S_j(x)=i$ le fait que la source $S_j$ attribue la classe $C_i$ à l'observation $x$. Nous supposons ici que les classes $C_i$ sont exclusives. A chaque source nous associons la fonction indicatrice~:
\begin{eqnarray}
M_i^j(x)=\left\{
    \begin{array}{l}
      1 \quad \mbox{si} \quad S_j(x)=i,\\
      0 \quad \mbox{sinon}.
    \end{array}
  \right.
\end{eqnarray}

La combinaison des sources s'écrit par~:
\begin{eqnarray}
\label{votesimple}
M_k^E(x)=\sum_{j=1}^{m}M_k^j(x),
\end{eqnarray}
pour tout $k$. L'opérateur de combinaison est donc associatif et commutatif. La règle du vote majoritaire consiste à choisir la décision prise par le maximum de sources, c'est-à-dire le maximum de $M_k^E$. Cependant cette règle simple n'admet pas toujours de solutions dans l'ensemble des classes $D=\{C_1,\ldots,C_n\}$. En effet, par exemple si le nombre de sources $m$ est paire et que $m / 2$ sources décident $C_{i_1}$ et $m/ 2$ autres sources disent $C_{i_2}$, ou encore dans le cas où chaque source affecte à $x$ une classe différente. Nous sommes donc obligé d'ajouter une classe $C_{n+1}$ qui représente l'incertitude totale liée au conflit des sources sous l'hypothèse de l'exhaustivité des classes $C_{n+1}=\{C_1,\ldots,C_n\}$. La décision finale de l'expert prise par cette règle s'écrit donc par~: 
\begin{eqnarray}
E(x)=\left\{
    \begin{array}{l}
      k \quad \mbox{si} \quad \max_k M_k^E(x),\\
      n+1 \quad \mbox{sinon.}
    \end{array}
  \right.
\end{eqnarray}
Cette règle est cependant peu satisfaisante dans les cas où deux sources donnent le maximum pour des classes différentes. La règle la plus employée est la règle du vote majoritaire absolu qui s'écrit~:
\begin{eqnarray}
E(x)=\left\{
    \begin{array}{l}
      k \quad \mbox{si} \quad \max_k M_k^E(x) >  \frac{m}{2},\\
      n+1 \quad \mbox{sinon.}
    \end{array}
  \right.
\end{eqnarray}

A partir de cette règle il a été démontré (Lam et Suen 1997) plusieurs résultats prouvant que la méthode du vote permet d'obtenir de meilleurs performances que toutes les sources prises séparément, sous des hypothèses d'indépendance statistique des sources et de même probabilité, et ceci est d'autant plus vrai que $m$ est impaire.

Il est possible de généraliser le principe du vote majoritaire afin de
supprimer le conflit. Au lieu de combiner les réponses des sources par une somme simple comme dans l'équation (\ref{votesimple}), l'idée est d'employer une somme pondérée (Xu et al. 1992)~:
\begin{eqnarray}
\label{voteponderegen}
M_k^E(x)=\sum_{j=1}^{m}\alpha_{jk} M_k^j(x),
\end{eqnarray}
où $\displaystyle \sum_{j=1}^{m}\sum_{k=1}^{n} \alpha_{jk}=1$. Les poids $\alpha_{jk}$ représentent la fiabilité d'une source pour une décision donnée, et l'estimation de ces poids peut se faire à partir des taux normalisés de réussite pour chaque classe et chaque classifieur. Notons qu'alors nous introduisons une connaissance {\it a priori} non nécessaire précédemment. Les différentes règles de décision possibles peuvent se résumer par la formule suivante~:
\begin{eqnarray}
\label{votegen}
E(x)=\left\{
    \begin{array}{l}
      k \quad \mbox{si} \quad M_k^E(x)=\max_i M_i^E(x)\geq
      c \, m+b(x),\\
      n+1 \quad \mbox{sinon,}
    \end{array}
  \right.
\end{eqnarray}
où $c$ est une constante de $[0,1]$ et $b(x)$ est une fonction de $M_k^E(x)$.

\subsection{Théorie des possibilités}
La théorie des possibilités proposée par (Zadeh 1978, Dubois et Prade 1988)
permet de tenir compte de l'imprécision des données ainsi que de
l'incertitude à partir de deux fonctions de possibilité et de
nécessité. Ces deux fonctions sont obtenues à partir des distributions
de possibilités définies sur $D=\{C_1,\ldots,C_n\}$ par~:
\begin{eqnarray}
\pi \, :\, D \rightarrow [0,1], \, \sup_{x\in D} \pi(x)=1. 
\end{eqnarray}
Ces distributions donnent le degré d'appartenance au domaine $D$, qui n'est autre qu'un opérateur flou. Afin d'extraire l'imprécision et l'incertitude des données, deux fonctions spécifiques sont définies à partir de ces distributions. La fonction de possibilité est définie pour tout $A \in 2^D$ par~:
\begin{eqnarray}
\Pi(A)=\sup_{x \in A}\pi(x).
\end{eqnarray}
La fonction de nécessité est donnée pour tout  $A \in 2^D$ par~:
\begin{eqnarray}
N(A)=1-\Pi(A^c),
\end{eqnarray}
où $A^c$ représente l'évènement contraire de $A$. 

Un des avantages de la théorie des possibilités est le nombre
d'opérateurs de combinaison disponibles. Il est ainsi possible de
combiner l'information issue des distributions de possibilité, à
partir d'opérateurs de type $t$-norme, $t$-conorme, moyenne, sommes symétriques, etc... Le choix du type de combinaison est un problème délicat {\it a priori} dans la théorie des possibilités, et doit être fait selon l'application et l'objectif recherché. Ce choix peut  se faire selon le comportement général de l'opérateur (conjonctif, disjonctif, ou des compromis), selon les propriétés désirées, selon sa capacité à discriminer les classes, ou encore selon son comportement dans des situations de conflit. En pratique de nombreux opérateurs sont employés et testés dans les applications, tels que $\max$ (opérateur de type $t$-norme), $\min$ (opérateur de type $t$-conorme), ou la moyenne, la médiane et les intégrales floues (opérateurs de type moyenne).

La dernière étape de la fusion d'informations est l'étape de décision. Dans le cadre de la théorie des possibilités, elle est généralement faite selon la règle suivante~: la classe $C_k$ est décidée pour l'observation $x$ si~:
\begin{eqnarray}
C_k=\argmax_{1\leq i \leq n} \mu_i(x),
\end{eqnarray}
où $\mu_i(x)$ représente le coefficient d'appartenance de $x$ à la classe $C_i$, qui sera ici donné par les sorties du classifieur.

Par construction des opérateurs de combinaison et de la règle de décision, la théorie des possibilités est davantage adaptée à la fusion d'informations de type numérique. Ainsi les coefficients d'appartenance peuvent être facilement obtenus dans le cadre de la classification par les sorties numériques des classifieurs. Nous emploierons donc ici cette théorie pour la fusion d'informations haut niveau sur des données de type numérique.

\subsection{Théorie des croyances}
La théorie des croyances (ou théorie de Dempster-Shafer) permet également de représenter à la fois l'imprécision et l'incertitude au travers deux fonctions~: la fonction de croyance et la fonction de plausibilité (Bloch 2003, Appriou 2002). Ces deux fonctions sont dérivées des fonctions de masses.  Le principe de la théorie des croyances repose sur la manipulation de ces fonctions de masse définies sur des sous-ensembles et non sur des singletons comme dans la théorie des probabilités. En effet, elles sont définies sur chaque sous-espace de l'ensemble des disjonctions du cadre de discernement $D=\{C_1,\ldots,C_n\}$ et à valeurs dans $[0,1]$. Généralement, il est ajouté une condition donnée par~:
\begin{eqnarray}
\label{hyp1_fonction_masse}
\sum_{A \in 2^D} m_j(A)=1,
\end{eqnarray}
où $m(.)$ représente la fonction de masse. Dans cette théorie la première difficulté est le choix de la fonction de masse. Plusieurs approches ont été proposées, nous en retenons ici deux~: l'une fondée sur un modèle probabiliste (Appriou 2002) et l'autre sur une transformation en distance (Den{\oe}ux 1995).

(Appriou 2002) propose deux modèles répondant à trois axiomes qui impliquent la considération de $n*m$ fonctions de masse aux seuls éléments focaux possibles $\{C_i\}$, $\{C_i^c\}$ et $D$. Un axiome garantit de plus l'équivalence avec l'approche bayésienne dans le cas où la réalité est parfaitement connue (méthode optimale dans ce cas). Ces deux modèles sont sensiblement équivalents sur nos données, nous utilisons dans cet article le modèle donné par~:
\begin{eqnarray}
\left\{
\begin{array}{l}
m_{ij} (C_i)(x)= \frac{\alpha_{ij}R_j p(S_j/C_i)}{1+R_j p(S_j/C_i)}\\
m_{ij} (C_i^c)(x)=\frac{\alpha_{ij}R_j}{1+R_j p(S_j/C_i)}\\
m_{ij}(D)(x)=1-\alpha_{ij} \\
\end{array}
\right.
\end{eqnarray}
où $p$ est une probabilité, $R_j =\left( \max_{i,j}p(S_j/C_i)\right)^{-1}$ est un facteur de normalisation, et $\alpha_{ij} \in [0,1]$ est un coefficient d'affaiblissement permettant de tenir compte de la fiabilité d'une source $S_j$ pour une classe $C_i$, que nous choisissons ici égale à 1.

La difficulté de ce modèle est alors l'estimation des probabilités $p(S_j/C_i)$. Dans le cas où la donnée de la source $S_j$ est la réponse d'un classifieur exprimée sous la forme de la classe (donnée symbolique), l'estimation de ces probabilités peut être faite par les matrices de confusion sur une base d'apprentissage. Si la réponse du classifieur est une donnée numérique, l'estimation de telles probabilités peut se faire soit par une approche fondée sur les fréquences, soit sous l'hypothèse de la distribution suivie par ces probabilités. Dans ce dernier cas l'estimation est généralement plus délicate, nous retiendrons donc ce modèle pour la fusion d'informations haut niveau des données symboliques. 

En revanche l'approche fondée sur une transformation en distance proposée par (Den{\oe}ux 1995) est plus adaptée à la fusion d'informations haut niveau des données numériques. En effet, les fonctions de masse sont définies par~:
\begin{eqnarray}
\left\{
\begin{array}{l}
	m_{ij}(C_i/x^{(t)})(x)=\alpha_{ij}\varphi_i\left(d(x,x^{(t)})\right)\\
	m_{ij}(D/x^{(t)})(x)=1-\alpha_{ij}\varphi_i\left(d(x,x^{(t)})\right)\\
\end{array}
\right.
\end{eqnarray}
où $\left(x^{(t)}\right)$ est un vecteur d'apprentissage des réponses des sources, $\alpha_{ij} \in [0,1]$ est un coefficient d'affaiblissement, $d$ est une distance à déterminer entre $x$ et $x^{(t)}$, $C_i$ est la classe associée à $x^{(t)}$, et $\varphi_i$ est une fonction vérifiant~:
\begin{eqnarray}
\left\{
\begin{array}{l}
	\varphi_i(0)=1,\\
	\displaystyle \lim_{d \rightarrow + \infty} \varphi_i(d)=0.\\
\end{array}
\right.
\end{eqnarray}
Il existe un grand nombre de fonctions $\varphi_i$ vérifiant ces égalités, sans qu'il y ait une méthode pour le choix de ces fonctions. Dans le cas d'une distance euclidienne, Den{\oe}ux propose la fonction~:
\begin{eqnarray}
\varphi_i(d)=\exp(\gamma_i d^2),
\end{eqnarray}
où $\gamma_i > 0$ est un paramètre associé à la classe $C_i$. $\gamma_i$ peut être initialisé comme l'inverse de la distance moyenne entre les vecteurs d'apprentissage vérifiant $C_i$. La distance $d(x,x^{(t)})$ peut être considérée uniquement pour les $k$-plus proches voisins de $x$ afin de réduire le temps de calcul. 

La différence de fond avec les modèles d'Appriou est qu'ici il faut estimer une distance au lieu d'une probabilité. La distance, généralement euclidienne est plus adaptée aux données numériques, tandis que l'estimation des probabilités est ici plus aisée pour des données symboliques. Dans le cas d'une distance euclidienne, notons que nous obtenons une fonction de masse proche de celle obtenue par un modèle d'Appriou sous l'hypothèse d'une distribution gaussienne. Nous comparons donc ces deux approches pour la fusion d'informations avec des données de type symétrique pour le modèle probabiliste et numérique pour le modèle des distances.

La combinaison des fonctions de masse est fondée sur la règle orthogonale de Dempster-Shafer non normalisée proposée par (Smets 1990) définie pour deux fonctions de masse $m_1$ et $m_2$ et pour tout $A \in 2^D$ par~:
\begin{eqnarray}
m(A)=(m_1\oplus m_2)(A)=\sum_{B\cap C =A} m_1(B)m_2(C).
\end{eqnarray}
Cette opérateur est associatif et commutatif. La masse affectée sur l'ensemble vide s'interprète comme une mesure de conflit. Pour les modèles d'Appriou et de Den{\oe}ux, nous obtenons donc une unique fonction de masse en combinant les fonctions $m_{ij}$.

La dernière étape de la fusion est la décision sur la classe la plus vraisemblable. La théorie des croyances offrent plusieurs règles de décision fondées sur la maximisation d'un critère. En particulier, nous pouvons employer le maximum des fonctions de croyance ou des fonctions de plausibilité. Si les premières peuvent être trop pessimistes, les secondes peuvent être trop optimistes. Un compromis est le maximum des probabilités pignistiques proposées par (Smets 1990) qui est le critère retenu dans cet article.

\section{Classification d'images sonar}
La classification des images sonar est un problème difficile en soi. Les méthodes de caractérisation automatique consistent en des méthodes d'analyse de texture, les images de fonds marins présentant en effet des zones de sédiments homogènes ou non qui peuvent s'apparenter à des textures. La littérature concernant les méthodes d'analyse de la texture est abondante et le choix de l'une ou de plusieurs d'entre elles dépend très souvent de l'image et de l'application. Ces méthodes fournissent généralement un ensemble assez restreint de paramètres pertinents qui permettent de classer l'image en un ou plusieurs type de sédiments. Nous exposons ici toute la complexité des images sonar due aux nombreuses imperfections, puis nous présentons les classifieurs à base de méthodes d'analyse de texture.

Les images sonar sont obtenues à partir des mesures temporelles faites en traînant à l'arrière d'un bateau un sonar qui peut être latéral, frontal, ou multifaisceaux. Chaque signal émis est réfléchi sur le fond puis reçu sur l'antenne du sonar avec un décalage et une intensité variable. Pour la reconstruction sous forme d'images un grand nombre de données physiques (géométrie du dispositif, coordonnées du bateau, mouvements du sonar, ...) est pris en compte, mais elles sont entachées des bruits de mesures dues à l'instrumentation. A ceci viennent s'ajouter des interférences dues à des trajets multiples du signal (sur le fond ou la surface), à des bruits de chatoiement, ou encore à la faune et à la flore. Les images sont donc entachées d'un grand nombres d'imperfections telles que l'imprécision et l'incertitude.

26 images fournies par le GESMA (Groupe d'Etudes Sous-Marine de l'Atlantique) ont été obtenues à partir d'un sonar Klein 5400 permettant une bonne résolution. Ces images ont été segmentées en imagettes de taille $64\times64$ pixels (voir Tab. \ref{tab_images_sonar}) étiquetées selon le type de sédiment. 
\begin{table}[!ht]
\begin{center}
  \begin{tabular}{lr}
  \begin{tabular}{c}
  \epsfig{file=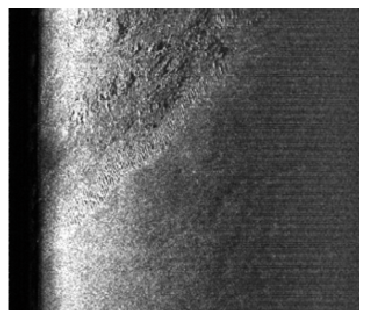, width=5cm}
  \end{tabular}
&
	\begin{tabular}{|c|c|c|}
  \hline
  & & \\
\epsfig{file=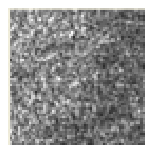, width=1cm}&
\epsfig{file=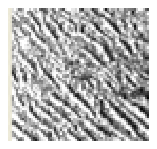, width=1cm}&
\epsfig{file=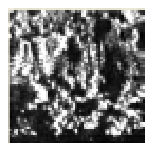, width=1cm}\\
  Sable & Ride & Roche\\
  \hline
  & & \\
\epsfig{file=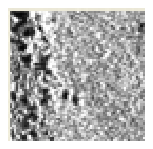, width=1cm}&
\epsfig{file=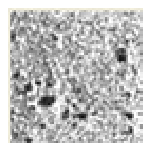, width=1cm}&
\epsfig{file=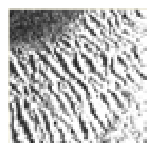, width=1cm}\\
  Sable   & Cailloutis  & Ride \\
  et Roche  &   & et Sable\\
  \hline
\end{tabular}
\end{tabular}
\end{center}
\caption{Exemple d'image sonar (fournit par le GESMA) et d'imagettes extraites et étiquetées.}
\label{tab_images_sonar}
\end{table}
Nous avons ainsi distingué le sable (54.52\%), la roche (21.35\%), les rides de sable (8.80\%), la vase (5.50\%), les cailloutis (0.77\%) et l'ombre (2.40\%) qui représente l'absence d'information sur le type de sédiment. De plus, nous avons indiqué lorsque ces imagettes comprennent plus d'un sédiment (homogènes ou non), ce qui représente 39.70\% des imagettes. Le type de sédiment de ces imagettes est le plus présent. Notons également que ces bases de données sont très délicates à réaliser, car elles sont entachées des erreurs éventuelles de l'expert.

Les classifieurs sont composés chacun d'une méthode d'extraction de paramètres de texture et d'un perceptron multicouche. L'approche retenue pour l'architecture de fusion est celle présentée sur la Fig. 1. La fusion d'informations haut niveau se fait donc soit au niveau des sorties numériques des perceptrons soit au niveau des sorties symboliques représentant les classes affectées.
\begin{figure}[h]
\begin{center}
\includegraphics[width=10cm]{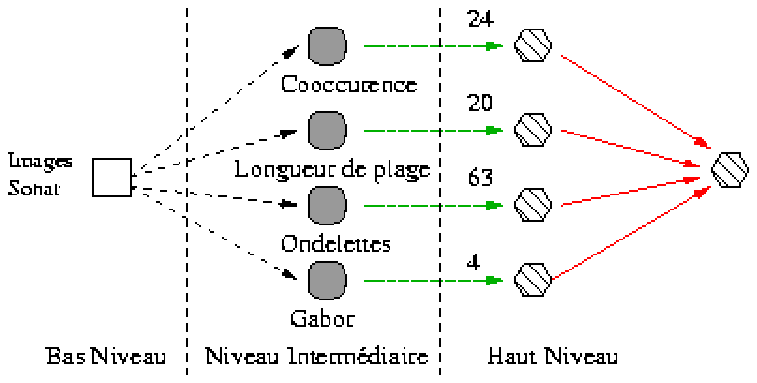}
\end{center}
\vspace*{-0.5cm}
\caption{Architecture de fusion de classifieurs retenue.}
\label{architecture}
\end{figure}

Les méthodes d'extraction de paramètres de texture sont celles déjà présentées dans (Martin et al. 2004). Chaque méthode permet de calculer des paramètres différents, parfois redondants entre eux, mais avec des caractéristiques propres à la méthode.

Les matrices de co-occurrence sont calculées en comptant les occurrences identiques de niveaux de gris entre deux pixels contigus. Quatre directions sont considérées~: 0, 45, 90 et 135 degrés. Dans ces quatre directions six paramètres d'Haralick sont calculés~: l'homogénéité, le contraste, l'entropie, la corrélation et l'uniformité. Un des problèmes principaux de cette approche est la non invariance en translation. Ainsi les imagettes de rides auront des paramètres différents selon la direction de celles-ci.

Les matrices de longueurs de plages sont obtenues en comptabilisant les pixels consécutifs possédant le même niveau de gris dans les quatre directions précédemment considérées. Dans chacune des directions cinq paramètres sont calculés~: la proportion de petite longueur de plage, la dispersion des plages entre les niveaux de gris et entre les longueurs et le pourcentage des longueurs de plage. Cette approche est bien adaptée aux images optiques faiblement bruitées. Dans le cas des images sonar, où un bruit de chatoiement est fortement présent, il faudrait soit supprimer ce bruit soit adapter le calcul des paramètres. Nous conservons cependant cette approche afin d'évaluer la robustesse de la fusion aux mauvais paramètres de texture.

La troisième approche retenue est une transformée en ondelettes. Les deux approches précédentes ne permettent pas de tenir compte de l'invariance dans les directions. La transformée en ondelettes discrète invariante en translation est fondée sur le choix de la transformation optimale pour chaque niveau de décomposition. Chaque niveau de décomposition fournit quatre images, sur lesquelles nous calculons trois paramètres~: l'énergie, l'entropie, et une  moyenne. Nous retenons un niveau de décomposition de trois ce qui fournit 63 paramètres au classifieur. 

Enfin, une approche fondée sur les filtres de Gabor permet de résoudre le problème des rides. En effet, nous considérons cinq fréquences différentes pour six directions ce qui donne trente filtres. Sur ces filtres, nous calculons quatre paramètres statistiques~: le maximum de l'écart-type d'un sédiment considéré, la moyenne de tous les filtres, la moyenne dans la direction horizontale (celle des pings du sonar) et l'écart-type avant filtrage.

Ces quatre jeux de paramètres sont ensuite considérés indépendamment à l'entrée de quatre perceptrons multicouche ayant ainsi des couches d'entrée de 24, 63, 20 et 4 neurones et une couche de sortie de 6 neurones correspondant aux six classes de sédiments considérés. L'apprentissage est réalisé pour une fonction sigmoïde de sortie donnant ainsi pour chacun des neurones $k$ de la couche de sortie une valeur réelle $o_k \in [0,1]$. Ces valeurs $o_k$ constituent les données numériques sur la décision des classifieurs. Les décisions symboliques sont obtenues en considérant pour chaque classifieur le maximum des $o_k$, indiquant ainsi la classe $C_k$ préférée par chaque perceptron.

\section{Résultats}
La base de données a été divisée aléatoirement en trois parties égales. La première sert à l'apprentissage des perceptrons multicouche, la deuxième à l'apprentissage de la fusion et la troisième pour les tests. Afin d'accroître la qualité de l'estimation des taux de classification, nous avons répété le tirage aléatoire 10 fois et moyenné les résultats. Dans l'approche par vote majoritaire nous avons obtenu un conflit de 18.59\%, afin de supprimer ce conflit
nous avons considéré l'approche avec les pondérations $\alpha_{jk}$ estimées par les matrices de confusion. Dans le cadre de la théorie des possibilités de nombreux opérateurs de combinaison ont été testés ; nous présentons ici celui donnant les résultats les plus probants donnés par l'opérateur $\max$ ($t$-conorme). 

Le Tab. 2 présente les taux de bonne classification définis par le rapport du nombre d'imagettes bien classées sur le nombre total d'imagettes de la base de test. 
\begin{table}[t]
\begin{center}
		\begin{tabular}{|c|c|c|c|c|c|c|c|c|}
		\hline
		Coocc. & longueur & ondel. & Gabor & PMC & vote & poss. &\multicolumn{2}{c|} {croyances} \\
		\cline{8-9}
		 & de plages &  &  &  & &  &  proba. & distance \\
		\hline
		70.0 & 50.3 & 68.9 & 66.4 &50.0 & 62.0 & 69.9 & 68.8 &79.5 \\
		\hline
		\end{tabular}
\end{center}
\caption{Taux de classification avant et après fusion d'informations (\%).}
\label{res_tot}
\end{table}
Nous constatons que les quatre méthodes de fusion présentées sont plus robustes aux données erronées fournies par les longueurs de plages que le PMC (perceptron global prenant en entrée l'ensemble des paramètres extraits). Cependant la fusion par vote reste moins bonne que les trois classifieurs issus des matrices de co-occurence, ondelettes et
Gabor, les hypothèses de (Lam et Suen 1997) ne sont pas vérifiées. Les deux approches de fusion d'informations haut niveau à partir des données numériques sont plus performantes que les  approches à partir des données symboliques. Notons de plus que la théorie des croyances avec le modèle de distance donne significativement les meilleurs résultats que nous détaillons dans le Tab. 3. 
\begin{table}[t]
\begin{center}
		\begin{tabular}{|c|c||c|c|}
		\hline
		roche & 87.3 & sable & 84.9\\
		\hline
		ride & 61.3 & vase & 4.9 \\
		\hline
		cailloutis & 0.9 & ombre & 71.5\\
		\hline
		homogènes & 91.3 & non homogènes & 63.1\\
		\hline
		\end{tabular}
\end{center}
\caption{taux de classification par type de sédiment pour le modèle de
  distance (\%).}
\label{res_detail_dist}
\end{table}
Les meilleurs taux sont atteints pour les sédiments sable et roche,
ceci est dû à l'apprentissage du perceptron qui est meilleur pour les
sédiments les plus représentés numériquement. Les cailloutis et la
vase offrent de mauvais résultats car leur effectif est faible dans la base. Notons encore que les taux pour les imagettes homogènes sont bien meilleurs, mais est-il raisonnable de chercher à affecter un type de sédiment à une imagette qui en contient plusieurs. Ceci doit entraîner une remise en cause de la constitution même de la base de données.

\section{Conclusion}
Nous avons étudié les différentes approches de fusion d'informations haut niveau, en faisant ressortir leurs avantages et inconvénients, et notamment la facilité pour chacune d'entre elles à être employées pour des données numériques et symboliques. Ces approches ont été comparées dans le cadre d'une application particulièrement délicate~: la classification d'images sonar. En effet, nous avons vu la complexité pour l'expert à interpréter ces images, et la difficulté de les
classer automatiquement. La fusion d'informations apporte une solution intéressante pour la résolution de tels problèmes particulièrement grâce à sa facilité de mise en {\oe}uvre pour des applications de classification. Nous avons ici fait ressortir de meilleures performances pour la fusion d'informations haut niveau à partir de données numériques et plus particulièrement dans le cadre de la théorie des croyances. Cependant, nous devons bien nous garder de généraliser de tels résultats à tout type de données. 

Pour cette application nous avons fait ressortir l'influence du sur-apprentissage du perceptron employé qui provient de la différence d'effectifs des sédiments dans la base de données. La gestion des évènements rares peut être réalisée par la fusion, mais dans ce cas avant le classifieur. Une fusion d'informations bas niveau doit alors être envisagée. Une autre difficulté est issue de la constitution même de la base. Le fait d'avoir des imagettes possédant plusieurs sédiments augmente l'incertitude, qui est dans ce cas dure à mesurer. Nous travaillons sur la réalisation d'une base de zones homogènes où l'incertitude sera mesurable plus finement.

\section*{Références}

\begin{list}{~}{\setlength{\labelsep}{0 cm}
		\setlength{\leftmargin}{0.6 cm}
		\setlength{\rightmargin}{0 cm}}

\item \hspace{-0.6 cm} Appriou, A. (2002), Discrimination multisignal par la théorie de l'évidence, In Décision et Reconnaissance des formes en signal, Hermes Science Publication, pp~219-258, 2002.  

\item \hspace{-0.6 cm} Bloch, I. (2003), Fusion d'informations en traitement du signal et des images, Lavoisier (eds), Hermes Science Publication, 2003.

\item \hspace{-0.6 cm} Dasarathy, B.V. (1997), Sensor Fusion Potential Exploitation - Innovative Architechtures and Illustrative Applications, Proceeding of the IEEE 1997, 85(1), pp~24-38. 

\item \hspace{-0.6 cm} Den{\oe}ux, T. (1995), A $k$-Nearest Neighbor Classification Rule Based on Dempster-Shafer Theory, IEEE Transactions on Systems, Man Cybernetics 1995, 25(5), pp~804-813.

\item \hspace{-0.6 cm} Dubois, D. et Prade, H. (1988), Possibility Theory, Plenum Press, New York, 1988.

\item \hspace{-0.6 cm} Lam, L. et Suen, C.Y. (1997), Application of Majority Voting to Pattern Recognition: An Analysis of Its Behavior and Performance, IEEE Transactions on Systems, Man Cybernetics 1997, 27(5), pp~553-568.

\item \hspace{-0.6 cm} Martin, A., Sévellec, G. et Leblond, I. (2004), Characteristics vs decision fusion for sea-bottom characterization, Caractérisation in-situ des fonds marins, Brest, France, 2004.

\item \hspace{-0.6 cm} Smets, Ph. (1990), The Combinaison of Evidence in the Transferable Belief Model, IEEE Transactions on Pattern Analysis and Machine Intelligence 1990, 12(5), pp~447-458.

\item \hspace{-0.6 cm} Xu, B.V., Krzyzak, A. et Suen, C.Y. (1992), Methods of Combining Multiple Classifiers and Their Application to Handwriting Recognition, IEEE Transactions on Systems, Man Cybernetics 1992, 22(3), pp~418-435.

\item \hspace{-0.6 cm} Zadeh, L.A. (1978),	Fuzzy Sets as a Basis For a Theory of Possibility, Fuzzy Sets and Systems 1978, 1, pp~3-28.

\end{list}

\section*{Summary}
In this paper, we present some high level information fusion approaches for numeric and symbolic data. We study the interest of such method particularly for classifier fusion. A comparative study is made in a context of sea bed characterization from sonar images. The classification of kind of sediment is a difficult problem because of the data complexity. We compare high level information fusion and give the obtained performance.
\end{document}